\documentclass[sigconf]{acmart}

\usepackage{soul,color}

\usepackage{accents}

\usepackage{algorithmicx} %
\usepackage{algpseudocode} %

\usepackage{booktabs}
\usepackage{wrapfig}
\usepackage{float}
\usepackage{mathtools}
\usepackage{graphicx,tipa}%

\usepackage{enumitem}

\acmConference[UrbComp'24]{The 13th International Workshop on Urban Computing
}{August, 2024}{Barcelona, Spain}

\renewcommand\footnotetextcopyrightpermission[1]{}
\usepackage{graphicx}
\usepackage{amsmath,amsfonts}
\usepackage{subfigure}
\usepackage{textcomp}

\usepackage{color,soul}
\sethlcolor{red}
\newcommand{\name}{{\tt BikeMAN}}

\begin{document}
\title{Micromobility Flow Prediction: A Bike Sharing Station-level 
 Study via Multi-level Spatial-Temporal Attention Neural Network
}
\author{Xi Yang \quad
Jiachen Wang \quad 
 Song Han \quad 
Suining He
}
\affiliation{
School of Computing, 
University of Connecticut
\country{USA}}

\begin{abstract}
Efficient use of urban micromobility resources such as bike sharing  is challenging due to the unbalanced station-level demand and supply, which causes the maintenance of the bike sharing systems painstaking. Prior efforts have been made on accurate prediction of bike traffics, i.e., demand/pick-up and return/drop-off,  to achieve system efficiency. 
However, bike station-level
traffic
prediction is difficult because of the spatial-temporal complexity
of bike sharing systems. Moreover, such level of prediction over entire bike sharing systems is also challenging due to the large number of bike stations. To fill this gap, we propose \name{}, a multi-level spatio-temporal attention neural network 
to predict station-level bike traffic for entire bike sharing systems. The proposed network consists of an encoder and a decoder with an attention mechanism representing the spatial correlation between features of bike stations in the system and another attention mechanism describing the temporal characteristic of bike station traffic. Through experimental study on over 10 millions trips
of bike sharing systems ($>$ 700 stations) of New York City, our network showed high accuracy in predicting the bike station traffic of all stations in the city.

\end{abstract}

\maketitle

\section{Introduction}

Bike sharing has become an essential micromobility option for urban residents worldwide due to its advantages in convenience and economy over other urban mobility means. Within a docked/station-based bike sharing system, 
a micromobility user picks up a bike from one station (demand), accomplishes her or his trip,  and drops the bike off at another station (return).

\begin{figure}[h]
  \centering
  \subfigure[Demands on Fri, Jun 7, 2019]{
  	\centering                                                     
  	\includegraphics[width = 0.4\columnwidth]{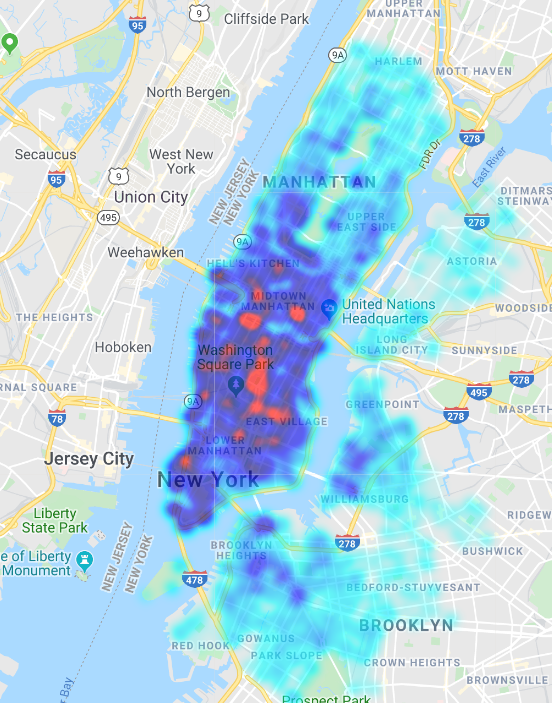}
  }
  \subfigure[Returns on Fri, Jun 7, 2019]{
  	\centering                                                     
  	\includegraphics[width = 0.4\columnwidth]{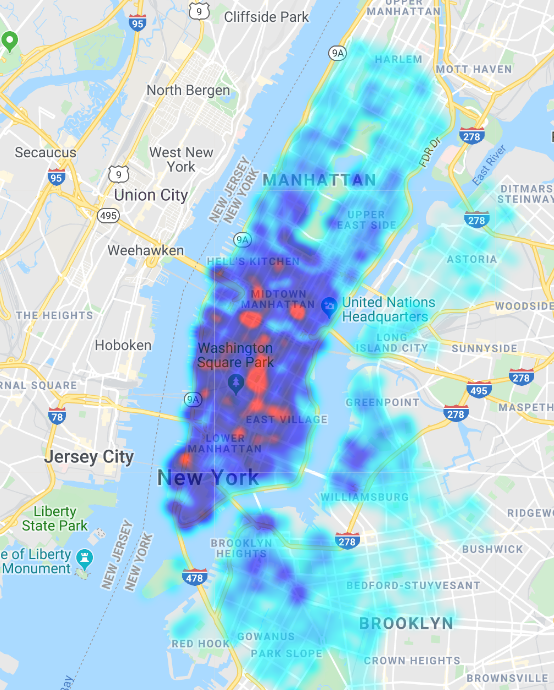}
  }
\caption{Imbalanced bike usage heatmap of NYC. Red areas represent high demand, while light blue areas represent low demand.}
	\label{fig_unbalance}
\end{figure}

Through data analysis of the bike sharing systems in several cities, a critical issue lies in that 
there were imbalanced demands and returns distributions in those bike sharing systems, i.e. a majority of demand and a majority of redistributed/returned bikes only occurred in a small portion of stations \cite{wang2019towards}. 
A typical imbalanced %
traffic map of the bike sharing system is shown in Figure \ref{fig_unbalance}, where the bike traffics spatially vary across the city map. 
Such imbalance not only degrades the user satisfaction, but also leads to operation inefficiency and resource waste to the bike sharing systems. 
How to predict the future bike traffic (pick-ups and drop-offs) and the subsequent rebalanced distribution of bikes in every station becomes essential and challenging. 
In order to solve the balancing problems, the general idea is propose an accurate pick-up and drop-off prediction model.

Bike sharing traffic (pick-up and drop-off) prediction can usually be defined as a time series prediction problem from multi-source and heterogeneous data \cite{li2019learning}. In this work we  propose a \underline{Bike} sharing \underline{M}ulti-level \underline{A}ttention neural \underline{N}etwork called \name{} for station-based pick-up and drop-off prediction with historical spatial and temporal traffic features and some external related factors (including weather conditions and points of interest (POIs)). Our work will not only give the prediction results for bike sharing system operators to efficiently rebalance their bike distribution, but also to provide information for citizens about the bike availability at the stations of their interest in future hours or minutes. Our main contributions are as follows:

\begin{itemize}
  \item[1)] \textit{Comprehensive bike data analysis}: We have provided a comprehensive and detailed real-world data analysis on how the weather conditions, station locations and surrounding POIs impact the bike usage in the metropolitan area like New York City, and visualized them to validate our insights. 
  \item[2)] \textit{Multi-level spatio-temporal attention model}:  To the best of our knowledge, we propose the \textit{first} novel multi-level attention neural network consisting of a spatial attention and a temporal attention mechanism to predict station-level bike usage (pick-ups and drop-offs), which accurately predicts the traffics of all the stations. 
  The spatial attention mechanism developed in this work captures not only the correlation of features within a single station, but also the correlation spatial of features across all stations in the city. The developed temporal attention mechanism captures temporal correlation between features across different timestamps.
  \item[3)] \textit{Extensive experimental studies}: We have conducted extensive experimental studies upon over 10 millions trips. Our multi-level attention neural network demonstrates high accuracy compared with other basic and encoder-decoder models in multi-station predictions.
\end{itemize}

The rest of the paper is organized as follows. 
We first review the related in work in Section \ref{sec:relatedwork}. 
After that, we present the data analysis in Section \ref{sec:dataanalysis}.
Given the features, we then present the core model of our approach in Section \ref{sec:model}.
We show the performance evaluation in Section \ref{sec:evaluation}, and conclude in Section \ref{sec:conclusion}.

\section{Related work}\label{sec:relatedwork}

The station-level prediction is challenging because of the spatial-temporal characteristics of the bike sharing demand pattern of a station and a large number of stations in a city. Recent studies have been focused on tackling this challenge. 
Hulot et al. \cite{hulot2018towards} proposed a statistical learning model using temporal and weather features to predict the hourly pick-ups and drop-off at each station of the system. Chen et al. \cite{chen2017prediction} used recurrent neural networks (\texttt{RNN}s) trying to capture the temporal characteristics in bike sharing demand. 
Li, et al.  \cite{li2019learning} proposed a representation learning method which encoded the spatial-temporal information of the bike sharing system, and the representation was then fed to Long-Short-Term-Memory (\texttt{LSTM}) to predict bike demand. Chai, et al. developed a multi-graph convolutional networks as the representation of the spatial-temporal data instead, but they also fed the learned representation to LSTM \cite{chai2018bike}. 
Lin et al. \cite{lin2018predicting} proposed a novel graph convolutional neural network with data-driven graph filter model to predict hourly demand for a large number of stations in the bike sharing systems. 
Despite the accuracy shown upon the selected bike stations (say, only 272 selected out of all $\sim$800 stations in \cite{lin2018predicting}), they have not extensively, systematically and comprehensively studied the \textit{entire} bike sharing system (like NYC) for station traffic prediction, and provided deployment insights like our work here. 

Attention mechanisms have been widely used in natural language processing \cite{bahdanau2014neural,luong2015effective}. The idea of implementing attention on neural networks extend to speech recognition \cite{chorowski2015attention,bahdanau2016end,zeyer2018improved}, video summarization \cite{ji2019video} and captioning \cite{song2017hierarchical}, and human action recognition \cite{song2017end}. Liang, et al. employed a attention mechanisms in an encoder-decoder structure for station-level geo-sensory time series prediction \cite{liang2018geoman}. 
Different from the prior studies, based on extensive data analysis, we design a novel multi-level attention mechanism for bike sharing system.

\section{Data Analysis}\label{sec:dataanalysis}
We first overview the datasets and then present the data analysis of the bike sharing dataset. 

\subsection{Data Overview} \label{sec4.1}

We use three types of data sets in our project: the user trip data, the weather condition data and POIs data. Details of each are discussed in the following.

\begin{itemize}
	\item[1)] \textbf{User Trip Data}: 
	It contains information of every single trip including the trip's duration, start/end time , start/end stations and their longitude/latitude, and user information. The user trip data of NYC bike sharing system is provided on a monthly base with over 2 million records each month. In this study, we use the user trip data from June 2019 to October 2019 collected online\footnote{https://www.citibikenyc.com/system-data}, with a total of over 10 millions trips.
	
	\item[2)] \textbf{Weather Condition Data}: 
    Characterizing the hourly weather historical conditions, it contains hourly temperature, precipitation, wind speed, humidity and other 6 weather observations. 
    We adopt the hourly weather data of New York City from June 2019 to October 2019\footnote{https://www.wunderground.com/history/daily/us/ny/new-york-city/KLGA/}.
	
	\item[3)] \textbf{POI Data}: POIs data, obtained from NYC Open Data Portal\footnote{https://data.cityofnewyork.us/City-Government/Points-Of-Interest/rxuy-2muj}, includes the GPS coordinates, facility domain and some other details (say, facility names) of over 20 thousand POI facilities in New York City. There are totally 13 major types of POIs (e.g. residential, transportation and commercial etc.), each of which contains a number of minor types. 
	
\end{itemize}

\subsection{Analysis}
We present the following insights regarding effects of weather and POIs upon the station traffics based on the data and visualization. 

\subsubsection{Station Traffic and weather}

\begin{figure}[h]
  \centering
  \includegraphics[width = \columnwidth]{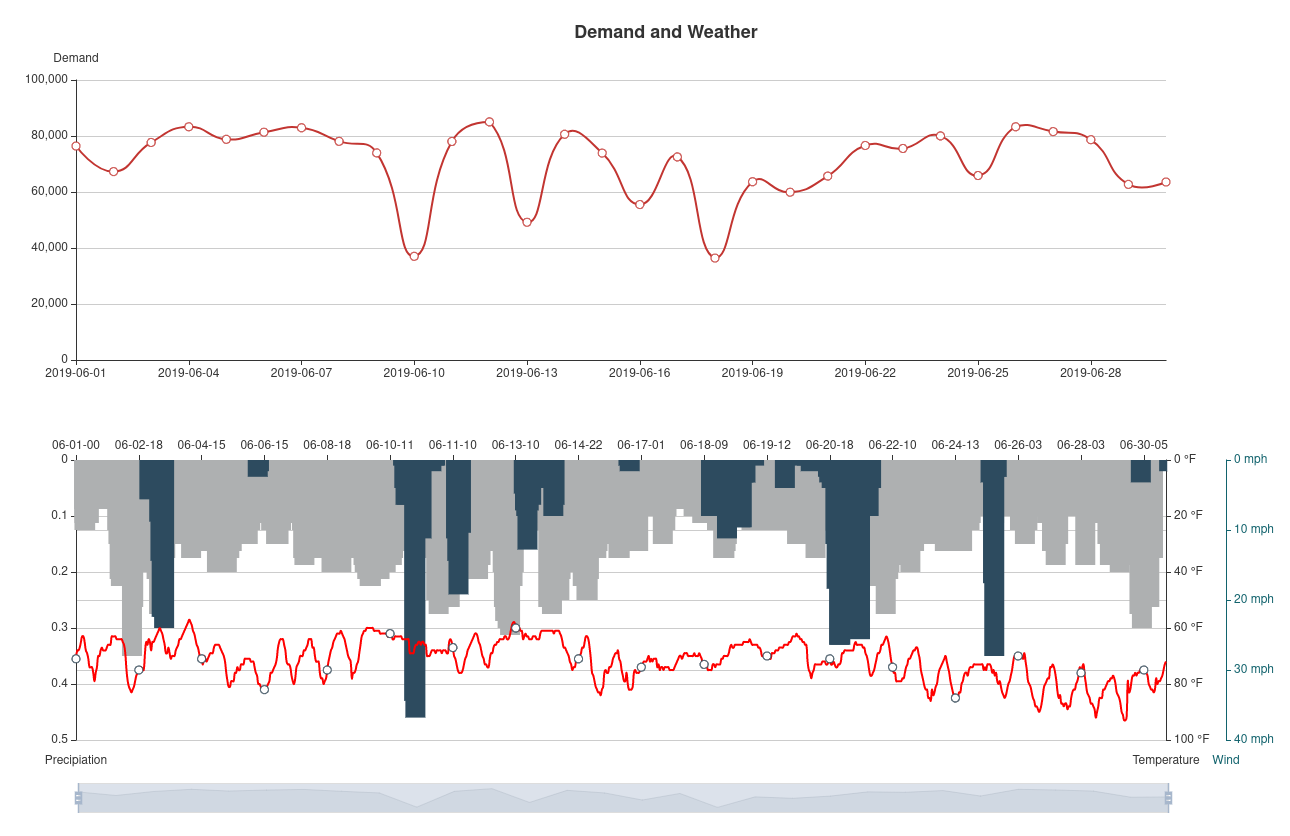}
  \caption{Bike demand and weather correlation in June, 2019. }
  \label{fig_weather}
\end{figure}

The bike usage is highly correlated with weather conditions. The two dominating effects are \textit{precipitation} and \textit{wind speed}. In general, wind speed and precipitation negatively impact the number of bike usage. Figure  \ref{fig_weather} shows the relationship between the hourly user trip amount of entire city and temperature, precipitation and wind speed in June, 2019 in NYC. Precipitation has the greatest influence on bike usage. On June 10, 0.46 inch of precipitation reduced the total amount of cycling from an average of 8000 to less than 4000. 

Wind speed is another important factor. Usually when the wind speed is higher than 15mph, a notable decrease in bike usage will happen. While the temperature's impact  on hourly bike demand is not as significant as the others, but its long-term influence cannot be ignored. Given above, we take those three factors into account and feed them to the deep learning model as input features.

\subsubsection{Station Traffic and POIs}
\begin{figure}[!ht]
	\centering
	\subfigure[Demand heatmap on Thursday, Jun 6, 2019]{
		\centering                                    
		\includegraphics[scale = 0.098]{6-7-demand}
	}
 \makebox[0.06in][]{}
	\subfigure[Distribution of commercial POIs]{
		\centering                                      
		\includegraphics[scale = 0.098]{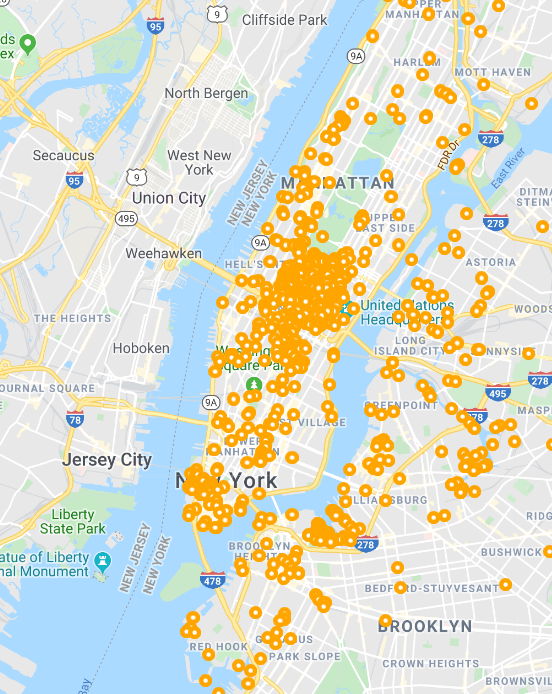}
	}
 \makebox[0.06in][]{}
	\subfigure[Distribution of residential POIs]{
	\centering                                                     
	\includegraphics[scale = 0.098]{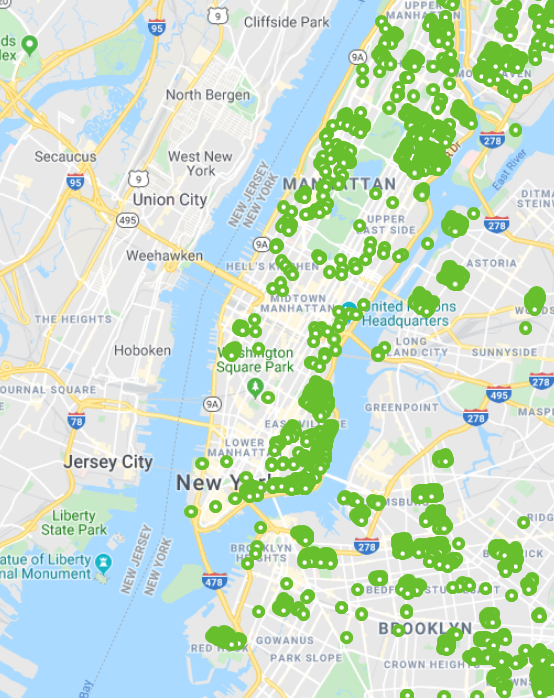}
	}
	\caption{Demand heatmap and distribution of commercial and residential POIs.}
	\label{fig_poi}
\end{figure}

From the previous demand heatmap, Figure  \ref{fig_unbalance}, we can also see that the bike usage of stations is highly location dependent. One of the major reasons behind this is that different locations have different types of POIs. Figure  \ref{fig_poi} illustrate this clearly. We use the distribution of commercial and residential POIs from the 13 categories as an example for visualization. The commercial POIs are mainly located on the center of Manhattan, surrounded by residential facilities. The different distributions of each types of POIs result in the dependency of bike usage on locations. Therefore, we take into account the distribution of the 13 types of POIs within $<$150m of each station as input features. 

In order to avoid the inclusion of another bike station when counting the nearby POIs around one station, we have to choose the distance to the station as small as possible. However, during our study, we found that the closest distance between two bike stations is less than 100m. If we use this closest distance, a large number of stations will have no POIs within this distance. Therefore, we set the distance of counting surrounding POIs to be 150 m. In this way, there are 91.25\% of stations of which distance between each other is larger than 150m, and there are only 6.92\% of stations having no POIs within this range, which suffices to characterize POI features. 

In summary, based on the analysis above, besides the bike demand we use weather conditions including precipitation, wind speed, and temperature, nearby ($< 150$ m) POIs, and station longitude and latitude as the input features of each bike stations.

\section{\name{} for Bike Station Traffic Prediction
}\label{sec:model}

\begin{figure}[h]
  \centering
  \includegraphics[width = \columnwidth]{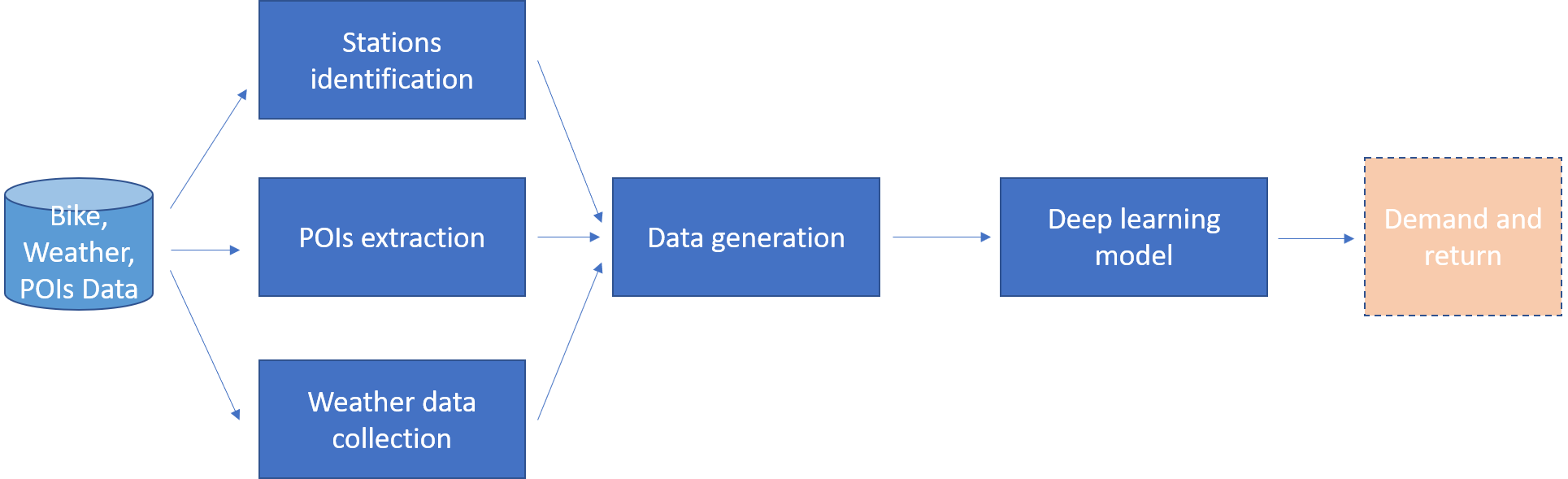}
  \caption{System architecture of \name{}. }
  \label{fig:sys}
\end{figure}

Our system's framework is shown in Figure~\ref{fig:sys}. Bike demand data set and two external features: hourly weather conditions data set and point of interest data set are processed respectively and then integrated together as the neural networks input. The model then will output the future demand prediction of one station.

\subsection{Preliminaries}
We consider that for totally \textit{N} stations in a city, each station \textit{i} has \textit{s} features at timestamp \textit{t} : \(\textbf{\textit{F}}_{i,t} = (F_{i1,t},...,F_{ij,t},...,F_{is,t})\), where \(i\in [1,...,N]\) and \(j\in [1,...,s]\), and let \(F_{i1,t}^{(p)}\) and \(F_{i1,t}^{(d)}\) be the bike traffic (pick-ups and drop-offs) of station \textit{i} at time \textit{t}.

The problem of this work is defined as:
given the features of each station of previous $\mathcal{T}$ timestamps, 
\begin{equation}
{\mathbf{\mathbb{F}}_{t-\mathcal{T}:t} = (\textbf{\textit{F}}_{1,{t-\mathcal{T}:t}},...,\textbf{\textit{F}}_{i,{t-\mathcal{T}:t}},...,\textbf{\textit{F}}_{N,{t-\mathcal{T}:t}})},   \quad i \in [1,...,N], 
\end{equation}
we aim at  predicting the bike traffic of all the stations $\hat{y}_{i,t:t+\tau}^{(P)}$ and $\hat{y}_{i,t:t+\tau}^{(D)}$, for all $i \in [1,...,N]$ at the next $\tau$ timestamps in the bike sharing system such that
 \begin{equation}
\min\sqrt{\sum_{k = 0}^\tau||\hat{y}_{i,t+k}^{(P)} - F_{i1,t+k}^{(P)}||^2}, \quad \text{and} \quad \min\sqrt{\sum_{k = 0}^\tau||\hat{y}_{i,t+k}^{(D)} - F_{i1,t+k}^{(D)}||^2}.
 \end{equation}

\subsection{Design of Multi-level Attention Neural Networks}
The structure of our network follows the encoder-decoder architecture with two attention mechanisms on the top of it. The \textit{encoder} and \textit{decoder} are two separate \texttt{RNN} components each of which is a stack of two \texttt{LSTM}s. In encoder part, a \textit{spatial attention} mechanism is implemented on the input features to generate the inputs of the \texttt{RNN} with the previous \texttt{RNN} states. In the decoder part, a \textit{temporal attention} mechanism is implemented on the output hidden states of the other \texttt{RNN} to compute the decoder outputs based on their correlation with encoder hidden states. 
We detail the two attention mechanisms as follows. 

\begin{figure}[h]
	\centering
	\includegraphics[width = \columnwidth]{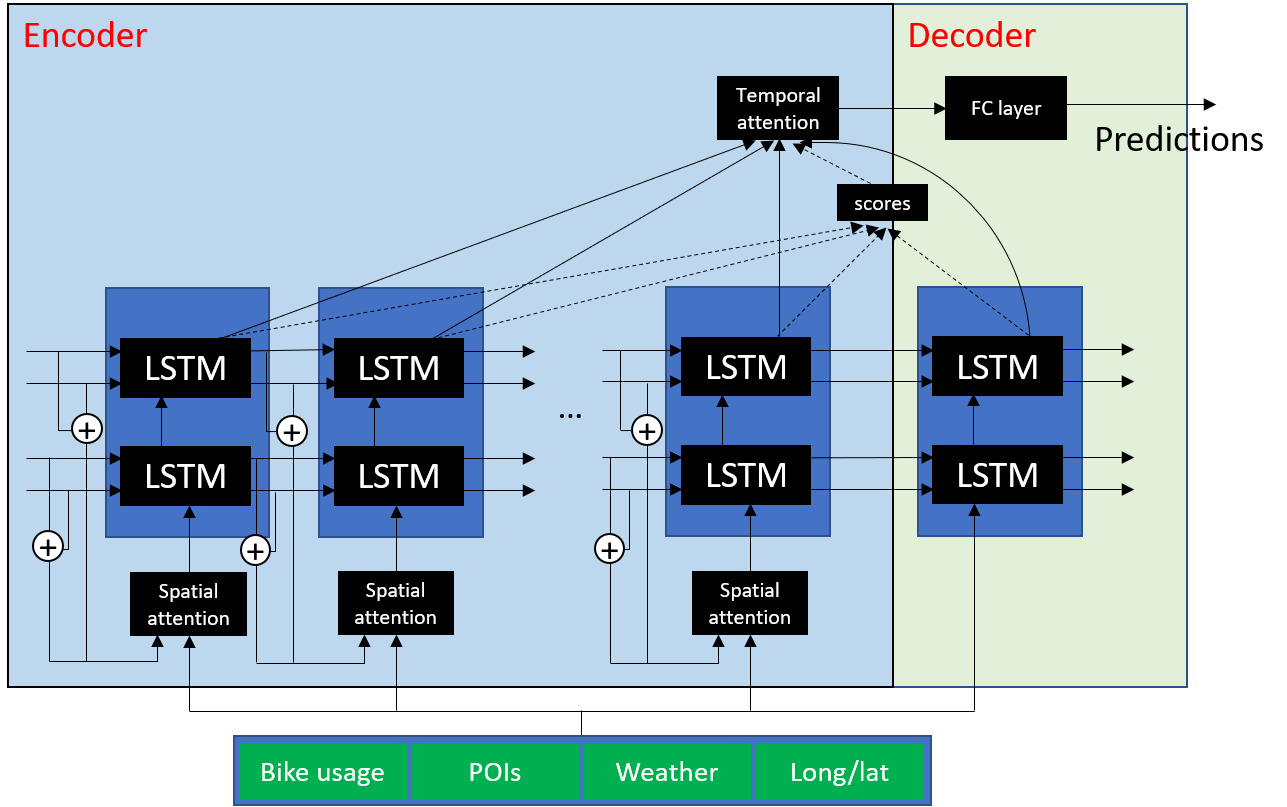}
	\caption{Core model architecture of \name.}
	\centering
	\label{fig1}
\end{figure}

\subsubsection{Spatial Attention}
The spatial attention mechanism is developed to describe the spatial correlation between the features of all stations. The intuition behind the spatial attention is that the traffic %
at each station is largely location-dependent as shown in Fig. \ref{fig_poi}, which can be characterized by the neighborhood POI distributions. 
For example, on a weekday morning demand in residential areas would be higher than commercial areas because people need to ride to work. Inspired by the work of \cite{liang2018geoman,song2017end,song2017hierarchical,bahdanau2016end,bahdanau2014neural,ji2019video}, we designed
the following spatial attention mechanism. 

Specifically, given the set of features of all the station of previous $\mathcal{T}$ timestamps, ${\textbf{$\mathbb{F}$}_{t-\mathcal{T}:t}}$, we first flatten it to create a vector $\mathbf{f}_{t-\mathcal{T}:t}$ so that the length at each timestamp is ${N\times s}$. For the $k$-th feature of the vector $\mathbf{f}_{t-\mathcal{T}:t}$, denoted as $\mathbf{f}_{t-\mathcal{T}:t}^{k}$, the score of this feature is calculated by Eq. (\ref{eq_sps}):
\begin{equation}
\epsilon^{k}_{t} = \mathbf{v}_{s}^\intercal \texttt{tanh}\left(\mathbf{W}_{s}[\mathbf{h}_{t-1};\mathbf{c}_{t-1}] + \mathbf{U}_{s}\mathbf{f}_{t-\mathcal{T}:t}^{k} + \mathbf{b}_{s}\right),
\label{eq_sps} 
\end{equation}
where $\mathbf{v}_{s}$, $\mathbf{b}_{s}$, $\mathbf{W}_{s}$, and $\mathbf{U}_{s}$ are the learnable model parameters,  $\mathbf{h}_{t-1}$ and $\mathbf{c}_{t-1}$ are the hidden state and cell state of the \texttt{RNN}, respectively, at the previous timestamp.
Then the attention weight of this feature is a \texttt{softmax} function of the score of this feature as in Eq. (\ref{eq_spw}):
\begin{equation}
\alpha^{k}_{t} = \frac{\texttt{exp}\left(\epsilon^{k}_{t}\right)}{\sum_{j = 1}^{N\times s}\texttt{exp}\left(\epsilon^{j}_{t}\right)},
\label{eq_spw} 
\end{equation} 
This way, the spatial attention captures the correlation between features of a station as well as the correlation between features across all of them. The input vectors of the \texttt{RNN} at timestamp $t$ is then computed by the multiplication of the attention weights and original input features of the model:

\begin{equation}
\Tilde{\mathbf{f}}_{t} = \left(\alpha^{1}_{t}\mathbf{f}_{t}^{1},\quad ..., \quad \alpha^{k}_{t}\mathbf{f}_{t}^{k},\quad ..., \quad \alpha^{N\times s}_{t}\mathbf{f}_{t}^{N\times s}\right).
\label{eq_spi} 
\end{equation}

\subsubsection{Temporal Attention}
The temporal attention mechanism captures correlation between features across different timestamps. The intuition behind this is that there is a strong time dependency of the demand of a station at time $t$ on previous demand. For example, on a sunny day morning if a large number of people choose to ride bikes to work, then in the evening the demand of stations around business areas is likely to be huge due to the return flows. The temporal attention scores between the current decoder \texttt{RNN} hidden state and one of the previous encoder \texttt{RNN} hidden states, $\lambda_{t,t'} $, are calculated as Eq. (\ref{eq_tmps}) by a concatenation manner as \cite{luong2015effective}:

\begin{equation}
\lambda_{t,t'} = \mathbf{v}_{l}^\intercal \texttt{tanh}\left(\mathbf{W}_{l}[\mathbf{h}_{t};\mathbf{h}_{t'}]\right),
\label{eq_tmps} 
\end{equation}
where $\mathbf{v}_{l}$ and $\mathbf{W}_{l}$ are learnable parameters, $\mathcal{T}$ is the number of timestamps of encoder, $\mathbf{h}_{t}$ is hidden state of encoder at timestamp $t$ which is in range of $[1,\mathcal{T}]$,  and $\mathbf{h}_{t'}$ is hidden state of decoder at timestamp $t'$. 
The attention weight, denoted as $\gamma_{t,t'}$, is then a \texttt{softmax} function of $\lambda_{t,t'} $ as in Eq. (\ref{eq_tmpw}):
\begin{equation}
\gamma_{t,t'} = \frac{\texttt{exp}\left(\lambda_{t,t'}\right)}{\sum_{t = 1}^{\mathcal{T}}\texttt{exp}\left(\lambda_{t,t'}\right)},
\label{eq_tmpw} 
\end{equation}
Finally, the output of the attention mechanism, $\mathbf{d}_{t'}$, is then a weighted sum of the encoder \texttt{RNN} hidden states shown in Eq. (\ref{eq_tmpo}):
\begin{equation}
\mathbf{d}_{t'}=\sum_{t=1}^{\mathcal{T}}\gamma_{t,t'} [\mathbf{h}_{t};\mathbf{h}_{t'}]. 
\label{eq_tmpo} 
\end{equation}

\subsubsection{Predictions and Model Training}
Through a fully connected (\texttt{fc}) layer, the concatenation of decoder \texttt{RNN} hidden state, $\mathbf{h}_{t'}$, and weighted sum of encoder \texttt{RNN} hidden states, $\mathbf{d}_{t'}$, is mapped to final predictions of traffics ($\hat{\mathbf{y}}_{t'} \in \{\mathbf{\hat{y}}_{t'}^{(P)}, \mathbf{\hat{y}}_{t'}^{(D)}\}$) at all stations of time $t'$: 

\begin{equation}
\hat{\mathbf{y}}_{t'} = \mathbf{W}_{p}[\mathbf{d}_{t'};\mathbf{h}_{t'}]+\mathbf{b}_p,
\label{eq_pred} 
\end{equation}
where $\mathbf{W}_{p}$ and $\mathbf{b}_{p}$ are learnable parameters.
The entire model is trained using the \texttt{Adam} optimizer which minimizes the mean square error between the predicted values and the ground truths as the loss function:
\begin{equation}
\mathcal{L}(\theta) = \sqrt{(\mathbf{y}_{t'}-\hat{\mathbf{y}}_{t'})^{2}}, 
\label{eq_loss}
\end{equation}
which is fast and efficient in practice. Based on the machines used, the training time for the New York City only takes 4 hours.

\section{Experimental Evaluations}\label{sec:evaluation}

We first present the evaluation setup, followed by the experimental results. 
\subsection{Evaluation Setup}
\subsubsection{Data Pre-processing}
We evaluate the \name{} and related schemes with the datasets presented in Section \ref{sec4.1}. All experimental evaluations are conducted upon a desktop with Intel i5-8700, 16GB RAM, an Nvidia GeForce GTX 1080Ti and Windows 10. We first go through all the bike sharing datasets and delete the stations which do not appear in all the datasets. We find 766 stations in common among all the datasets from Jun, 2019 to Oct, 2019. Then we extract the nearby ($< 150$ m) POIs around each of those common bike stations. The time interval is set to be 1 hour. The time stamps of encoder inputs are set 12, and the time stamps of decoder inputs are set 1. This means that we want to use this model to predict the return of all the stations in the next following hours based on their demand of previous 12 hours. 

Since some of the weather conditions have multiple data points between two continuous whole points, we take mean values of these data over an hour as the hourly data of this time. 
Most of the precipitation data is 0 as it is usually sunny in NYC. However, when it is rainy, the hourly precipitation is not high, typically within 1 inch, which is too small compared to the bike demand/return. 
Therefore, we normalize the precipitation data by min-max normalization such that they fall in the range of [0, 10]. For longitude and latitude data, the difference between two stations is even smaller, typically at the magnitude of $10^{-2}$. Therefore, we normalize these two features again by min-max normalization to make them fall in the range of [0, 100]. 
We use the trips in June, July, and August 2019 for training and validating our model, and use the ones in October, 2019 to test our trained model.

\subsubsection{Parameter Setups}
The parameters are set as followed. The learning rate is 0.001. The rate of gradients clipping is 2.5. The dropout rate is 0.3. The number of layers stacked in \texttt{LSTM}s is 2. The total number of stations is 766, each of which has 19 features. The number of encoder time stamps is 12 and that of decoder is 1. Since the output of \name{} is a vector of length 766 with each element being the demand for a station, the dimensions of hidden states $h$ in every \texttt{LSTM} of encoder and decoder are 1,024. The batch size is set to be 64. Total number of training epochs is 100 with 2,300 iterations.

\subsubsection{Comparison Metrics}
Root mean square error (RMSE) and mean absolute error (MAE) are selected as the metrics for evaluations:
\begin{equation}
    RMSE = \sqrt{\frac{1}{M}\sum_{i}^{M}(y_{i}-\hat{y}_{i})^{2}}, 
\label{eq5}
\end{equation}
\begin{equation}
    MAE = \frac{1}{M}\sum_{i}^{M}(y_{i}-\hat{y}_{i}),
\label{eq6}
\end{equation}
where \textit{M}, $y_{i}$ and $\hat{y}_{i}$ are the total number of predictions made by the model, the ground-truth of the bike pick-ups/drop-offs and the predicted bike pick-ups/drop-offs, respectively.

\subsection{Evaluation Results}
\subsubsection{Predictions by Varying Number of Involved Stations}
As a starting point, we first evaluate \name{} on bike traffic of a single station. 
We choose three stations for evaluation according to their demands sorted from high to low. 
Station 519 is the most popular bike station in NYC with the highest monthly traffic among all datasets. We then extend to carry out multi-station prediction, that is, training and predicting multiple station using a single model at the same time. 
Besides the 766 stations which we find commonly existing among all the datasets,  we choose here 190 stations out of the 766 stations that are common among all the datasets. 
The 190 stations we choose is based on the total demand from June to October 2019. We first sort all the stations by the total demand of and then pick 190 out of them that equally split the stations array. 
The accuracies of our model are compared with the encoder-decoder of the same structure as \name{} but no attention mechanisms on it. 
The comparison accuracies are shown in Table \ref{tb_nstation}.

\begin{table}[h]
\vspace{0.16in}
	\caption{Predictions of \name{} compared with basic encoder-decoder}
	\label{tb_nstation}
	\begin{center}
		\begin{tabular}{|l|ll|ll|}
			\hline
			\multicolumn{1}{|c|}{\bf Number of Stations}  &\multicolumn{2}{|c|}{\bf \name{}} &\multicolumn{2}{|c|}{\bf \tt LSTM Enc-Dec}\\
			\quad & RMSE & MAE & RMSE & MAE\\
			\hline
			1 (station 519)        & 11.70 & 7.29 & 11.78 & 7.16\\
			190       & 3.86 & 1.95 & 5.95 & 3.26\\
			766       & 3.79 & 2.08 & 5.64 & 3.26 \\
			\hline
			
		\end{tabular}
	\end{center}
\end{table}

The results are very interesting as shown in Table \ref{tb_nstation}. It turns out that the more stations we test, the more effective the attention mechanism is in improving the model accuracy. Only with multi-station predictions will \name{} outperform the basic encoder-decoder. The reasons probably lie in the spatial attention mechanism we design. In this work, the spatial attention is able to capture not only the correlation between the bike demand and other features for a single station, but also the correlation between the features of stations across the entire city. If there is only one station's features fed as input, the model will lose the information of others. Therefore, our model works better for predictions with multiple stations as inputs rather than single-station predicitons.

\subsubsection{Entire system predictions}
For operators of a bike sharing system, it is rather important to forecast the pick-ups and drop-offs for every station in the system than just a portion of stations. Therefore, we carry out extensive experimental test on the demand/pick-up and return/drop-off predictions for all 766 bike stations in NYC and compare our model with the baselines. We vary the \texttt{RNN} unit in \name{} from \texttt{LSTM} to Gated Recurrent Unit (GRU) (\name{}-{\tt GRU}). Our models are compared with basic \texttt{LSTM} and GRU encoder-decoder. The results of station traffic predictions are shown in Table \ref{tb_dmdrt}.

\begin{table}[h]
\vspace{0.16in}
	\caption{Predictions of \name{} compared with basic encoder-decoder}
	\label{tb_dmdrt}
	\begin{center}
		\begin{tabular}{|l|ll|ll|}
			\hline
			\multicolumn{1}{|c|}{\bf Models}  &\multicolumn{2}{|c|}{\bf Demand} &\multicolumn{2}{|c|}{\bf Return}\\
			\quad & RMSE & MAE & RMSE & MAE\\
			\hline
			{\tt GRU Dec-Enc}       & 5.661 & 3.312 & 5.729 & 3.386\\
			{\tt LSTM Dec-Enc}       & 5.678 & 3.350 & 5.693 & 3.338\\
			\name{}-{\tt GRU}       & 3.461 & 1.884 & 3.441 & 1.852\\
			\name{}       & 3.366 & 1.818 & 3.369 & 1.797 \\
			\hline
			
		\end{tabular}
	\end{center}
\end{table}

From Table \ref{tb_dmdrt}, we can see that our model \name{} with \texttt{LSTM} as the \texttt{RNN} units has the best performance among all the models for both demand and return predictions. The RMSE and MAE of \name{} drop by over 40\% from {\tt LSTM Dec-Enc}. 
This demonstrates the success of the spatial and temporal attention mechanisms proposed in Sec. \ref{sec:model}, which are able to capture more information in the heterogeneous input data. The model accuracy decreases slightly when \texttt{LSTM} cells are replaced by \texttt{GRU} cells. For each of the models tested, the prediction accuracy of demand is almost the same as that of return. 
From Fig. \ref{fig_unbalance} it can be seen that the demand and return have similar pattern over 24 hour period, which explains the similar performance of models on demand and return predictions. The predicted station traffics for the first week of October, 2019 by \name{} and \name{}-{\tt GRU} compared to the ground truth are shown in the figures below. We can see that \name{} highly resembles the ground-truths, validating the accuracy and effectiveness. 
\begin{figure}[h]
    \centering
    \includegraphics[scale=0.5]{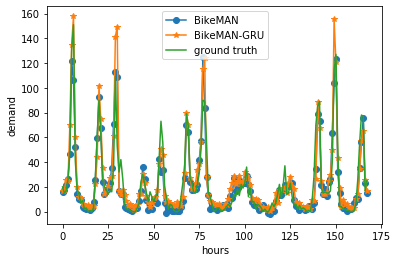}
    \caption{Prediction of \name{} and \name{}-{\tt GRU} on the bike demand of station 519 for the first week of October, 2019 (red line) compared with ground truth.}
    \label{fig6}
\end{figure}

\begin{figure}[h]
    \centering
    \includegraphics[scale=0.5]{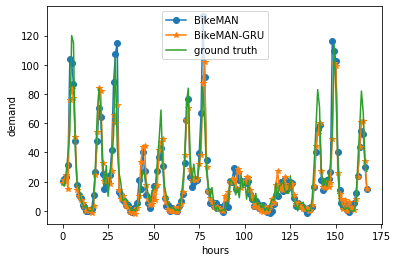}
    \caption{Prediction of \name{} and \name{}-{\tt GRU} on the bike return of station 519 for the first week of October, 2019 (red line) compared with ground truth.}
    \label{fig7}
\end{figure}

\section{Conclusion}\label{sec:conclusion}

In order to realize accurate and practical bike sharing station traffic prediction, we propose \name{}, a novel multi-level attention neural network. We design a novel encoder-decoder architecture, and leverage the spatial and temporal attentions to capture the correlations between the station features. \name{} further takes into account the external features like weather and points of interest to enhance the performance. We have conducted extensive data analysis and experimental studies upon the dataset from the New York City, demonstrating the accuracy of our proposed scheme.

\section{Acknowledgement}

This project is supported, in part, by 
the National Science Foundation (NSF) under Grant 2303575, 
the Google Research Scholar Award Program (2021--2022), 
the NVIDIA Applied Research Accelerator Program Award (2021--2022),
and the University of Connecticut (UConn) Research Excellence Program Awards (2020--2021, 2022--2024).

\bibliographystyle{ACM-Reference-Format}
\bibliography{bikeman}

\end{document}